\documentclass[letterpaper, 10 pt, conference]{ieeeconf}  

\usepackage[T1]{fontenc}  
\usepackage{cite}

\usepackage{amsmath}

\usepackage{comment}

\usepackage{graphicx}
\usepackage{subfigure}

\usepackage{xcolor}

\usepackage{hyperref}

\IEEEoverridecommandlockouts                              

\title{\LARGE \bf
Multi-UAV Uniform Sweep Coverage in Unknown Environments: A \textcolor{black}{Self-organizing Nervous System (SoNS)}-Based Random Exploration
}

\author{Aryo Jamshidpey$^{1}$ and Hugh H.-T. Liu$^{1}$
\thanks{$^{1}$Aryo Jamshidpey and Hugh H.-T. Liu are with the University of Toronto Institute for Aerospace Studies (UTIAS),
Toronto, Canada,    {\tt\small aryo.jamshidpey@utoronto.ca}
{\tt\small hugh.liu@utoronto.ca}
        }%
}

\begin{document}

\maketitle
\thispagestyle{empty}
\pagestyle{empty}

\begin{abstract}


\textcolor{black}{This paper addresses multi-UAV uniform sweep coverage in an unknown convex environment, where a homogeneous UAV swarm must evenly visit every portion of the environment for a sampling task without access to their position and orientation. Random walk exploration is practical in this scenario because it requires no localization and is easy to implement on swarms. We demonstrate that the Self-Organizing Nervous System (SoNS) framework, which enables a robot swarm to self-organize into a hierarchical ad-hoc communication network using local communication, is a promising control approach for random exploration in such environments. To this end, we propose a SoNS-based random walk method in which UAVs self-organize into a line formation and then perform a random walk to cover the environment while maintaining that formation. We evaluate our approach in simulations against several decentralized random walk strategies. Results show that our SoNS-based random walk achieves full coverage faster and with greater coverage uniformity than these benchmark strategies, both globally and in local regions.}

\end{abstract}

\section{INTRODUCTION}


Multi-UAV sweep coverage, where a group of UAVs collectively explore and cover an environment, has gained significant attention due to its importance in applications such as search and rescue~\cite{senanayake2016search}, wildfire monitoring~\cite{pham2017distributed,casbeer2006cooperative,lin2018topology}, and environmental surveillance~\cite{basilico2015deploying}. This paper investigates a uniform sweep coverage scenario in an unknown GNSS-denied convex environment, where a UAV swarm must collectively and evenly visit every portion of an unknown GNSS-denied \textcolor{black}{convex} environment---i.e., \textcolor{black}{a convex environment where UAVs have no prior knowledge of its shape, size, or whether it is convex or non-convex} and do not have access to global positioning data to determine their location or orientation---to perform uniform sampling.


Typically, Random Walk exploration is a practical strategy in unknown GNSS-denied environments because it is simple, scalable, flexible, and robust, requiring no localization or communication. Its ease of implementation makes it a suitable choice for robot swarms, but this approach can result in increased repeated coverage~\cite{kegeleirs2019random,dimidov2016random,tan2021comprehensive}. Incorporating cooperative methods can mitigate such redundancy and improve overall efficiency.

Building on our previous work\textcolor{black}{~\cite{jamshidpey2020multi, jamshidpey2024centralization}}, where we proposed the \textcolor{black}{Self-organizing Nervous System (SoNS)}~\cite{zhu2020formation,zhu2024self} as a \textcolor{black}{high-performing, scalable control approach with good fault tolerance against robot failures} for multi-robot sweep coverage with UGVs guided by UAVs, this paper applies this framework to a multi-UAV uniform sweep coverage task in an unknown GNSS-denied \textcolor{black}{convex} environment, focusing on random exploration using a homogeneous UAV swarm.


The \textcolor{black}{SoNS} is a formation control framework that enables a robot team to achieve a degree of central coordination without sacrificing the benefits of decentralized control through a self-organizing process. More specifically, under this framework an aerial-ground swarm can self-organize into an ad hoc communication network with a hierarchical structure using vision-based relative positioning and local communication. The UAV that occupies the highest level of the hierarchy through self-organization\textcolor{black}{, the "brain,"} is responsible for swarm-level decision-making and sending motion instructions downstream, while other robots use the network to cede authority to the brain and report sensing events upstream. By sending control information downstream, the robots establish a target formation and maintain it as the brain moves, facilitating effective exploration. \textcolor{black}{SoNS also provides practical on-the-fly capabilities such as collective sensing, merging, splitting, self-healing, and self-reconfiguring its control hierarchy, communication structure, and formation (e.g., to navigate narrow passages like corridors)~\cite{zhu2024self}.}

\textcolor{black}{This paper demonstrates} that the \textcolor{black}{SoNS} framework is an effective control approach for achieving multi-UAV uniform sweep coverage in unknown GNSS-denied \textcolor{black}{convex} environments. \textcolor{black}{We introduce} \textcolor{black}{a SoNS}-based random walk approach\textcolor{black}{, in which UAVs maintain a line formation and} employ a random walk strategy to cover the environment, and we \textcolor{black}{evaluate it in simulations} against four decentralized cooperative and non-cooperative random walk-based approaches as benchmarks. Additionally, we include an ideal SoNS-based approach in our comparison as a baseline, incorporating two additional assumptions. Our results indicate that the \textcolor{black}{SoNS}-based random walk achieves superior performance compared to the random walk-based methods, reducing the time to achieve full coverage and improving coverage uniformity across the entire environment and within local regions.

\section{RELATED WORK}

\textcolor{black}{Coverage in wireless sensor networks and robotics is typically classified into continuous and sweep coverage~\cite{elhabyan2019coverage,gorain2014approximation}. Continuous coverage involves stationary robots \textcolor{black}{continuously} monitoring an area, which becomes impractical for large environments due to the high number of robots required. In contrast, sweep coverage, also known as coverage path planning (CPP)~\cite{GALCERAN20131258,tan2021comprehensive,cabreira2019survey}, employs mobile robots to cover the environment. CPP methods are either offline, relying on prior knowledge to pre-plan paths~\cite{10421780,8894171}, or online, adapting to uncertain environments using sensors without any prior information~\cite{10102336,8286947}.}

Random walk is a widely recognized online sweep coverage technique, particularly in scenarios where robots lack localization, orientation, and prior knowledge of the environment~\cite{tan2021comprehensive}. \textcolor{black}{It involves agents
moving independently in random directions, randomly changing their direction at intervals or upon encountering boundaries or obstacles.} Various random walk strategies have been explored for exploration and coverage tasks~\cite{dimidov2016random,kegeleirs2019random}. In~\cite{kegeleirs2019random}, the authors compared five random walk strategies for swarm mapping in unknown environments, including Brownian motion~\cite{feynman1963mainly}, correlated random walk~\cite{renshaw1981correlated}, Lévy walk~\cite{zaburdaev2015levy}, Lévy taxis~\cite{pasternak2009levy}, and ballistic motion (equivalent to Random Billiards~\cite{comets2009billiards}). These methods are categorized in the literature as either fully random or specialized in coverage or exploration tasks. The study found that random billiards significantly outperforms 
other methods due to its higher coverage speed.

Using virtual pheromones is a common approach to reduce redundant coverage, based on which agents avoid revisiting recently explored areas\textcolor{black}{~\cite{dorigo1992optimization,rosalie2018chaos,ge2005complete}. In~\cite{kuiper2006mobility},} an efficient pheromone-based method for sweep coverage by UAVs was proposed, which has been adapted to our scenario. In another study~\cite{albani2017field}, a UAV swarm was employed for weed coverage and mapping in farmlands, utilizing a reinforced random walk~\cite{stevens1997aggregation}, and both explicit communication and pheromone-based cooperation. However, this approach is unsuitable for our GNSS-denied scenario because it relies on UAVs having knowledge of a predefined grid and position sharing. \textcolor{black}{Similarly, a connectivity-aware pheromone-based model~\cite{devaraju2023connectivity} requires UAVs to communicate their positions to maintain network connectivity, limiting its feasibility for GNSS-denied coverage.}

\section{PROBLEM STATEMENT}


This research addresses the problem of multi-UAV sweep coverage, where a homogeneous swarm of camera-equipped UAVs must collectively and evenly visit every portion of an unknown GNSS-denied environment to perform a uniform sampling task. The task, \textcolor{black}{predefined for the robots, is to take} one picture per each portion of the environment 
(with each portion representing a non-overlapping area of one square meter), requires UAVs to fly at a specific constant altitude (sampling altitude) and with a linear velocity not exceeding a target value, referred to as the target sampling velocity. While UAVs can fly at various velocities and altitudes, sampling is only valid when they adhere to these specific conditions.

In this research,  for simplicity, we represent the \textcolor{black}{convex} environment as a square arena, a common practice in the literature, conceptually decomposed into 1\,m\,x\,1\,m cells (i.e., the portions defined by the sampling task). This decomposition facilitates the performance evaluation of the studied methods; however, the UAVs are unaware of this grid and must operate without relying on it. A cell is considered visited only when a UAV enters it while performing the sampling task and adhering to the required speed and altitude conditions. Otherwise, the cell is not considered visited.



\section{METHODS}

In this section, we describe the evaluated approaches and the simulation setup. To ensure a fair comparison, all random walk-based strategies use Random Billiards as their core behavior due to its efficiency in exploration tasks~\cite{jamshidpey2024centralization,comets2009billiards}. This method involves agents moving in straight lines at a constant velocity until reflecting off a boundary in a random direction. Additionally, two benchmark cooperative approaches adapt swarm dispersion principles, a strategy commonly employed in continuous area coverage to maximize inter-agent distance and enhance the monitored area~\cite{bayert2019robotic,vijay2017received,beal2013superdiffusive,khaluf2018collective}, which we adapt here to reduce local UAV densities. To ensure fair comparison, all approach parameters were fine-tuned in an initial testing phase.

\subsection{SoNS-BASED METHODS}
\subsubsection{\textcolor{black}{SoNS}-BASED BOUSTROPHEDON SWEEP}
\textcolor{black}{The SoNS-based Boustrophedon Sweep (SoNS-BS) approach, serving as a baseline in our comparison, is adapted from \cite{jamshidpey2023reducing,jamshidpey2024centralization}.} The original approach enabled a heterogeneous swarm to self-organize into a hierarchical ad-hoc communication network with a caterpillar tree topology and a line formation, a common shape suitable for sweeping and mapping tasks~\cite{liu2018survey}. The brain UAV served as the root of the tree, ground robots acted as leaf nodes, and intermediate UAVs supervised subsets of ground robots while also being guided by another UAV.

\begin{figure}[t]
   \centering
      \includegraphics[width=3.4in]
      {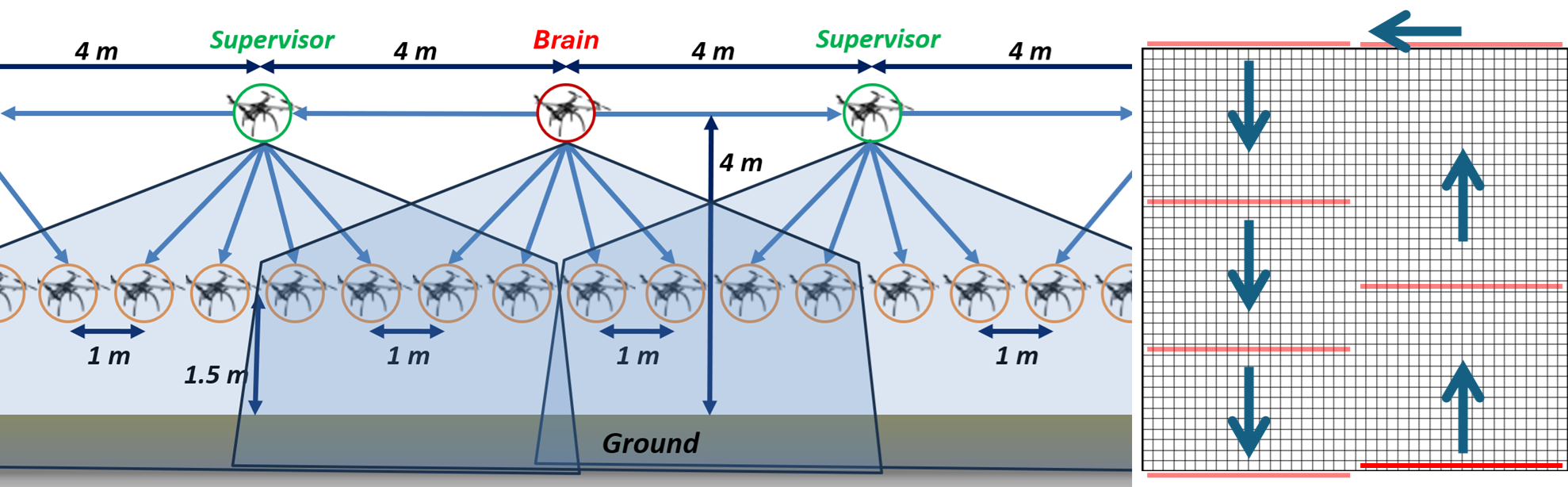}
   \caption{UAV positions in the formation and network topology for the \textcolor{black}{SoNS}-based approaches (left), and \textcolor{black}{SoNS}-BS sweep strategy (right).}
   \label{MNS_UAV_Line}
\end{figure}

\textcolor{black}{Here, we adapt the framework to a homogeneous UAV swarm, maintaining its self-organizing hierarchical structure. UAVs operate at two flight altitudes, referred to as supervisory and sampling. Using the same self-organizing process, they form a line formation for coverage (Fig.~\ref{MNS_UAV_Line}). The spacing between sampling UAVs is task-dependent; for instance, with 1\,m × 1\,m sampling cells, the spacing is set to 1\,m but SoNS can adjust it for different resolutions (e.g., 2\,m for 2\,m × 2\,m cells). This configuration minimizes gaps and overlap, ensuring no cell is missed between two sampling UAVs. Our setup consists of 25 UAVs: 5 supervisors (including the brain) and 20 sampling UAVs. Nevertheless, as shown in our previous work, SoNS is scalable and supports various setups. For details on the self-organization process, see~\cite{zhu2020formation}.} 

\textcolor{black}{After establishing the line formation, the brain UAV reactively sweeps the environment using standard back-and-forth boustrophedon motions~\cite{choset1998coverage} (Fig.~\ref{MNS_UAV_Line}) at the target sampling velocity.} 
Whenever the brain reaches a boundary, the SoNS exits the boundary by 50\,cm, shifts by the length of the formation, and begins a new sweep\textcolor{black}{. This process ensures full coverage, with each cell visited exactly once by the sampling UAVs, making the approach deterministic with a guaranteed completion time for a given convex environment
(for more details on the original approach, see~\cite{jamshidpey2024centralization}). For this approach to serve as an ideal baseline in our comparison, the robots must know that the environment is convex and begin by sweeping an outermost sub-region from its boundary, ensuring full coverage without revisiting any cell---two additional assumptions beyond the scenario.}

\subsubsection{\textcolor{black}{SoNS}-BASED RANDOM WALK}
In the \textcolor{black}{SoNS}-based Random Walk approach (\textcolor{black}{SoNS}-RW), UAVs first self-organize into the same hierarchical network with line formation used in \textcolor{black}{SoNS}-BS. Then, the brain UAV moves straight forward in a random direction at the target sampling velocity, followed by the other UAVs, until it reaches a boundary. As mentioned earlier, in the \textcolor{black}{SoNS} framework, the formation is consistently maintained throughout the brain's movement, including during the brain's rotations.

Upon reaching a boundary, the brain continues moving straight forward, crossing the boundary until it is more than 95\,cm away. At this point, it selects a random relative orientation, $\theta_{rand}$, directed towards the interior of the arena, excluding any orientation within a $30^\circ$ offset from its reciprocal heading. The brain then chooses the rotation direction, $D_{rand}$ (clockwise or counterclockwise), which requires the least time to reach the selected orientation.

If the rotation direction $D_{rand}$ 
is the same as the optimal rotation direction for aligning the formation along the boundary, denoted as $D_{adjust}$, the brain adjusts the formation accordingly. This adjustment is performed by rotating at an angular velocity that ensures no sampling UAV exceeds the target sampling velocity while maintaining the formation. During this phase, the UAVs continue to perform the sampling task as their velocities remain within the allowed range.

After either adjusting the formation along the boundary (when $D_{rand}$ 
is the same as $D_{adjust}$) or determining no adjustment is necessary (if $D_{rand}$ 
is not the same as $D_{adjust}$), the brain UAV rotates to the randomly selected orientation, $\theta_{rand}$. During this "Preparation" phase, the sampling task is temporarily paused, allowing UAVs to exceed the target sampling velocity to reposition effectively.

\textcolor{black}{After reaching $\theta_{rand}$, the brain resumes moving straight forward, continuing the random walk.}


\subsection{BENCHMARCK DECENTRALIZED METHODS}
\subsubsection{RANDOM BILLIARDS WITH RANDOMIZED OBSTACLE AVOIDANCE}
In this strategy, referred to as RB, each UAV moves forward with the target sampling velocity unless it encounters a boundary or obstacle. When a UAV is within 5\,cm of a boundary, it selects a random relative orientation directed towards the interior of the arena, avoiding directions within a $5^\circ$ offset from its reciprocal heading.

Obstacle avoidance has two priority levels: for short-range cases (higher priority, obstacles within 1\,m and $90^\circ$ ahead), the UAV randomly turns $10^\circ$--$30^\circ$, while for medium-range cases (lower priority, within 2.5\,m and $60^\circ$ ahead), it randomly turns $5^\circ$--$70^\circ$. The UAV turns clockwise if the obstacle is on the left and counterclockwise if on the right. After rotating, it resumes moving forward if no obstacles or boundaries are detected.

\subsubsection{LOCAL DENSITY REDUCTION THROUGH COMMUNICATION AND RANDOM REACTIONS}
In this strategy (LDR-Random), UAVs utilize RB as their core behavior and attempt to reduce local density using wireless communication with neighbors for a more uniform distribution in the arena. Each UAV broadcasts its ID within a 10\,m range at each step. If it receives at least five distinct IDs in a step, it detects high local density and notifies neighbors.

Upon receiving a high-density notification, the UAV randomly rotates $70^\circ$–$90^\circ$ clockwise\textcolor{black}{, then resumes RB and deactivates density reduction for 50 steps.} Additionally, after obstacle avoidance or boundary reactions, the UAV deactivates density reduction for 350 steps.

\subsubsection{LOCAL DENSITY REDUCTION THROUGH COMMUNICATION AND REPULSIVE REACTIONS}
In this approach (LDR-Repulsive), UAVs utilize RB as their core behavior, similar to LDR-Random. Additionally, this approach requires each UAV to accurately estimate the position of the sender of any received message within its local reference frame, enabling a more precise density reduction strategy.

Each UAV broadcasts its ID within a 5\,m range at every step. If it receives at least three distinct IDs within a step, it notifies its neighbors of high density.

Upon receiving such a \textcolor{black}{notification}, the UAV calculates the average position of detected neighbors and then determines the vector from its current position to that average position and the opposite direction. It then rotates via the shortest path to face that opposite direction. After completing such a rotation or an obstacle avoidance, the UAV resumes RB \textcolor{black}{and deactivates density reduction} for 350 steps.

\subsubsection{PHEROMONE-BASED MOBILITY MODEL}

This approach (PM), adapted from~\cite{kuiper2006mobility}, utilizes RB as its core behavior, similar to LDR-Random and LDR-Repulsive. Although artificial pheromones are challenging to implement in practice, PM serves as a benchmark for minimizing repeated coverage in sweep coverage tasks. 

Each UAV deposits 5000 units of virtual pheromone upon visiting a cell, which evaporates at a rate of 1 unit per step. UAVs read pheromone levels in their current and neighboring cells to guide movement decisions.
When no pheromone is detected ahead, UAVs move straight forward at the target sampling velocity. Otherwise, if the pheromone response is active, they probabilistically choose their next direction: moving forward with probability $p_A = \frac{\text{total} - \text{ahead}}{2 \times \text{total}}$, turning right with $p_R = \frac{\text{total} - \text{right}}{2 \times \text{total}}$, and turning left with $p_L = \frac{\text{total} - \text{left}}{2 \times \text{total}}$. Here, ``left'', ``right'', and ``ahead'' represent the pheromone levels detected in the corresponding directions, and ``total'' is their sum.

If a UAV turns left or right, it executes a $45^\circ$ turn before resuming forward motion. After a pheromone-based reaction, the pheromone response is deactivated for 25 steps. Additionally, after obstacle avoidance or boundary reactions, the pheromone response remains deactivated for 50 steps.

\subsection{SIMULATION SETUP}

The simulations were conducted in the ARGoS simulator \cite{pinciroli2012argos}, using UAV models implemented with an existing plugin \cite{allwright2018argos,allwright2018simulating}. The simulation took place in a 40\,m\,x\,40\,m square arena, centered at $(0, 0)$ in the coordinate frame\textcolor{black}{, where boundaries are detectable if within a UAV’s camera field of view.} For each approach, 30 simulation runs were conducted with 25 UAVs. 
The sampling altitude, supervisory altitude, and target sampling velocity were set to 1.5\,m, 4\,m, and 1\,m/s, respectively, with a communication range of up to 10\,m. For the \textcolor{black}{SoNS}-based approaches, the initial steps allow the formation to self-organize. Once the formation is established, the brain UAV is positioned at (10\,m,\,-19.5\,m,\,4\,m) for \textcolor{black}{SoNS}-BS (with the entire formation aligned along the southern boundary) and at (20\,m,\,-20\,m,\,4\,m) for \textcolor{black}{SoNS}-RW (above the southeastern corner), 
with the entire formation turned to match the brain UAV's randomly selected orientation toward the interior. 
For decentralized approaches, UAVs are randomly and uniformly placed within a  3\,m\,x\,20\,m area located inside the square arena along the middle of the southern boundary, at the 1.5 m flight altitude. UAVs start with a minimum spacing of 1.5\,m and random orientations between $0^\circ$ and $180^\circ$, all facing the interior. These initial steps are not recorded in the data. The dataset is available at \href{https://doi.org/10.5281/zenodo.13764325}{https://doi.org/10.5281/zenodo.13764325}. 



\section{SIMULATION RESULTS}
This section presents the simulation results, evaluating three metrics: coverage completion time (CCT), total coverage uniformity (TCU), and local coverage uniformity (LCU).

\subsection{COVERAGE COMPLETION TIME (CCT)}
\begin{figure}[t]
   \centering
      \includegraphics[height=1.8in,keepaspectratio]{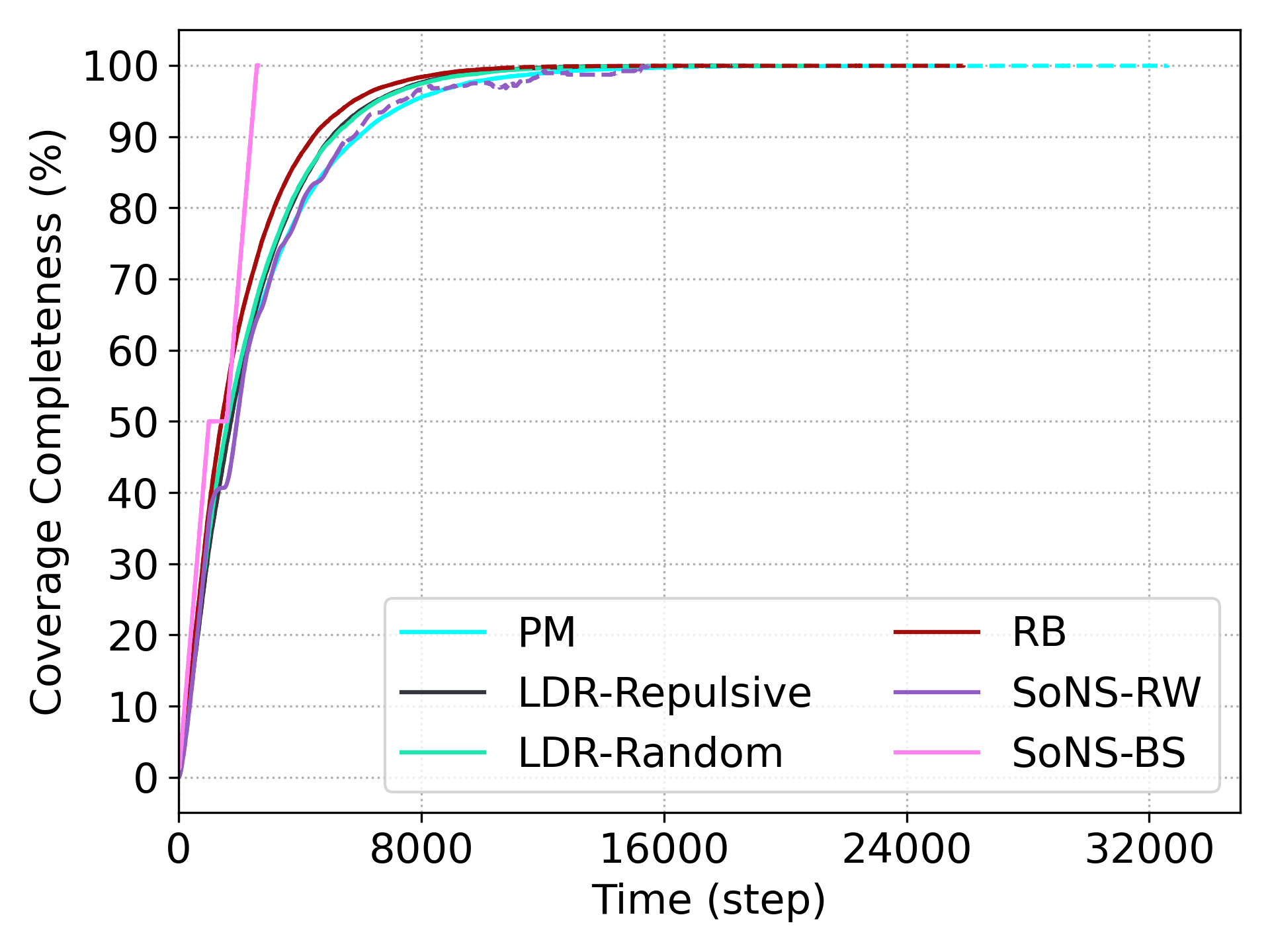}
   \caption{\textcolor{black}{Mean} coverage completeness of each approach over time.}
   \label{Fig:Time_Line}
\end{figure}

\begin{figure*}[t] 
    \centering
    \subfigure[Coverage completion time.]{%
        \includegraphics[width=0.32\textwidth]{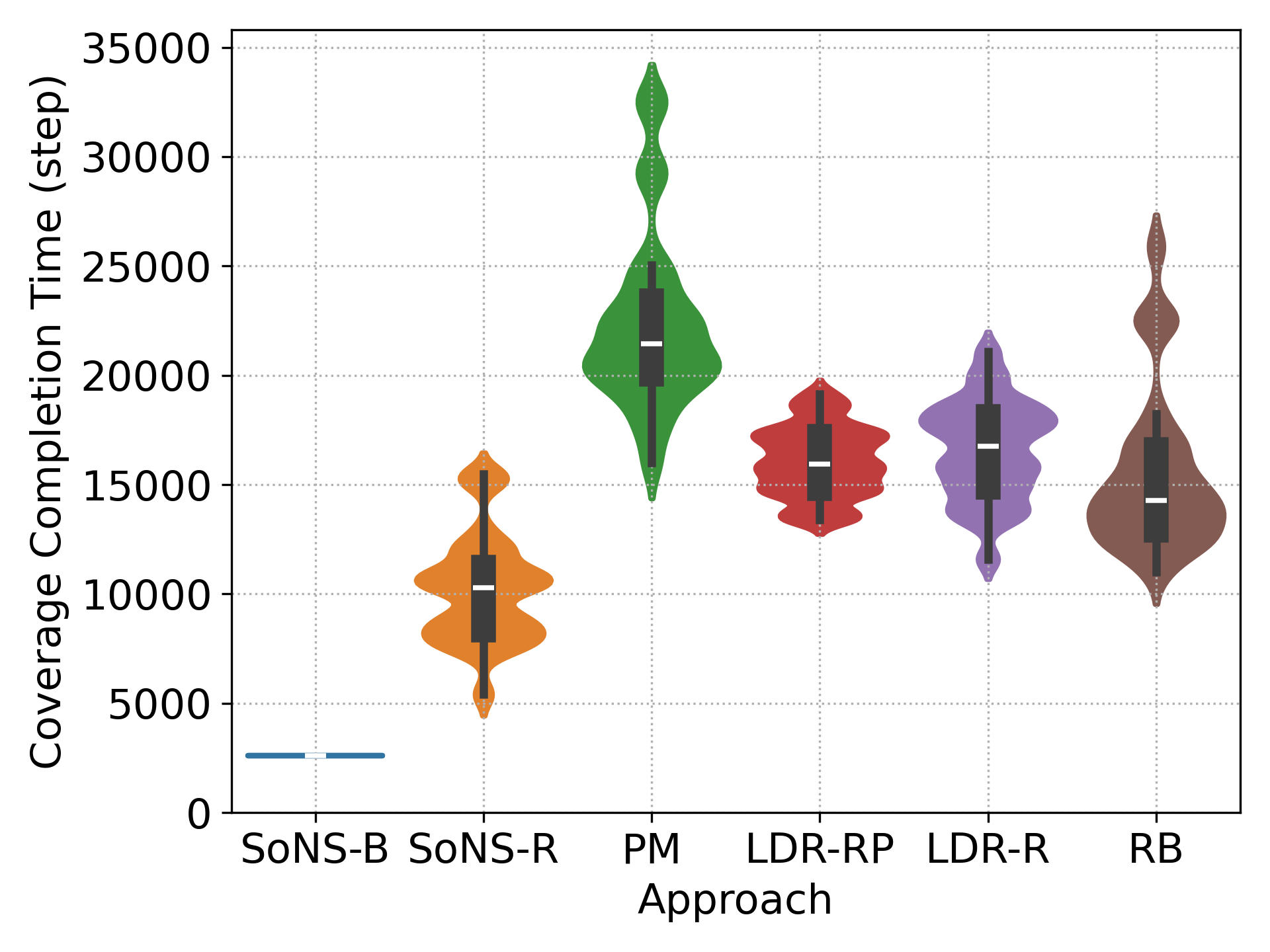}
        \label{Fig:Time_Violin}
    }
    \hfill
    \subfigure[Total uniformity at coverage completion.]{%
        \includegraphics[width=0.32\textwidth]{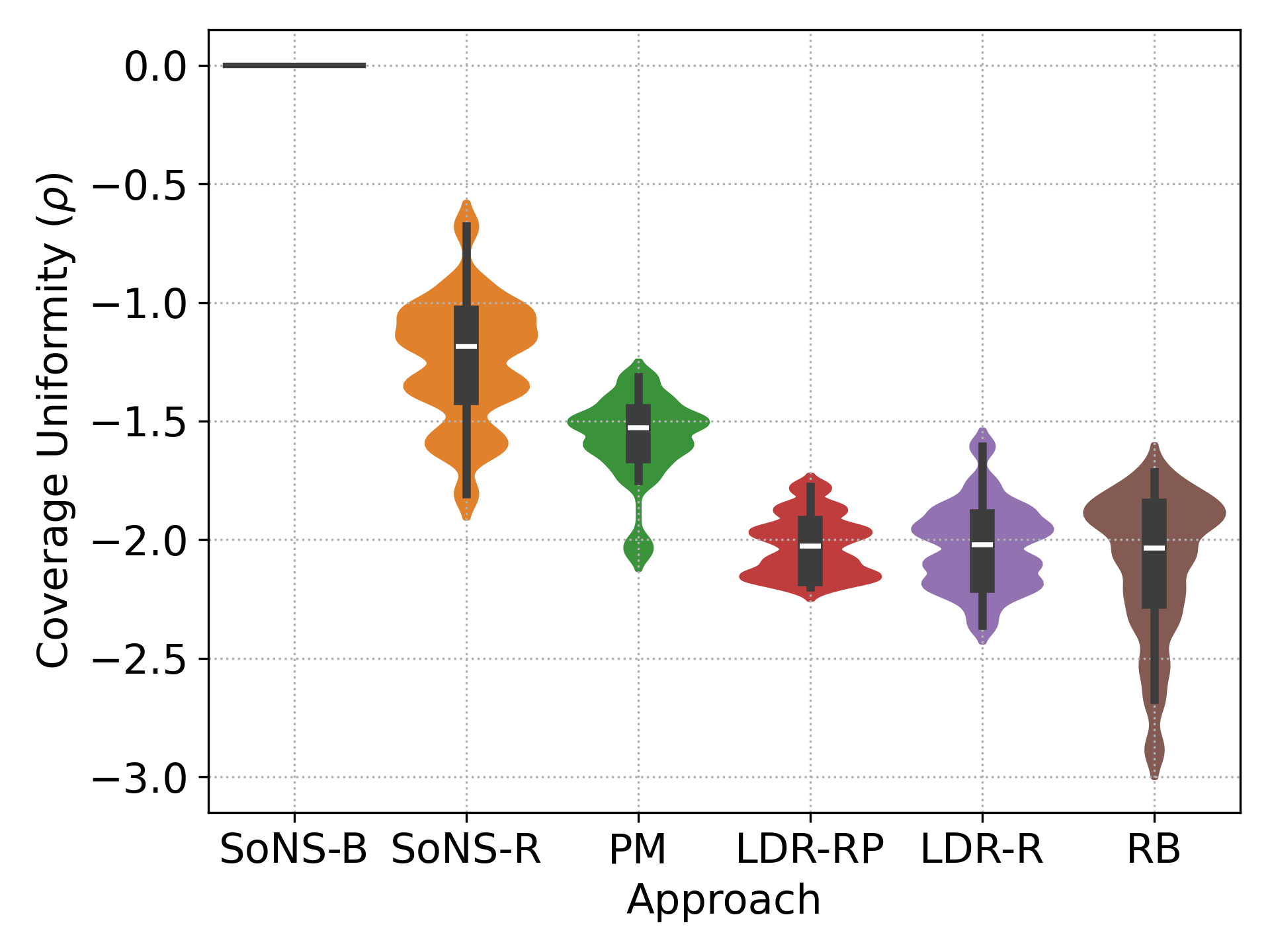}
        \label{Fig: Total Uniformity}
    }
    \hfill
    \subfigure[Local uniformity at coverage completion.]{%
        \includegraphics[width=0.32\textwidth]{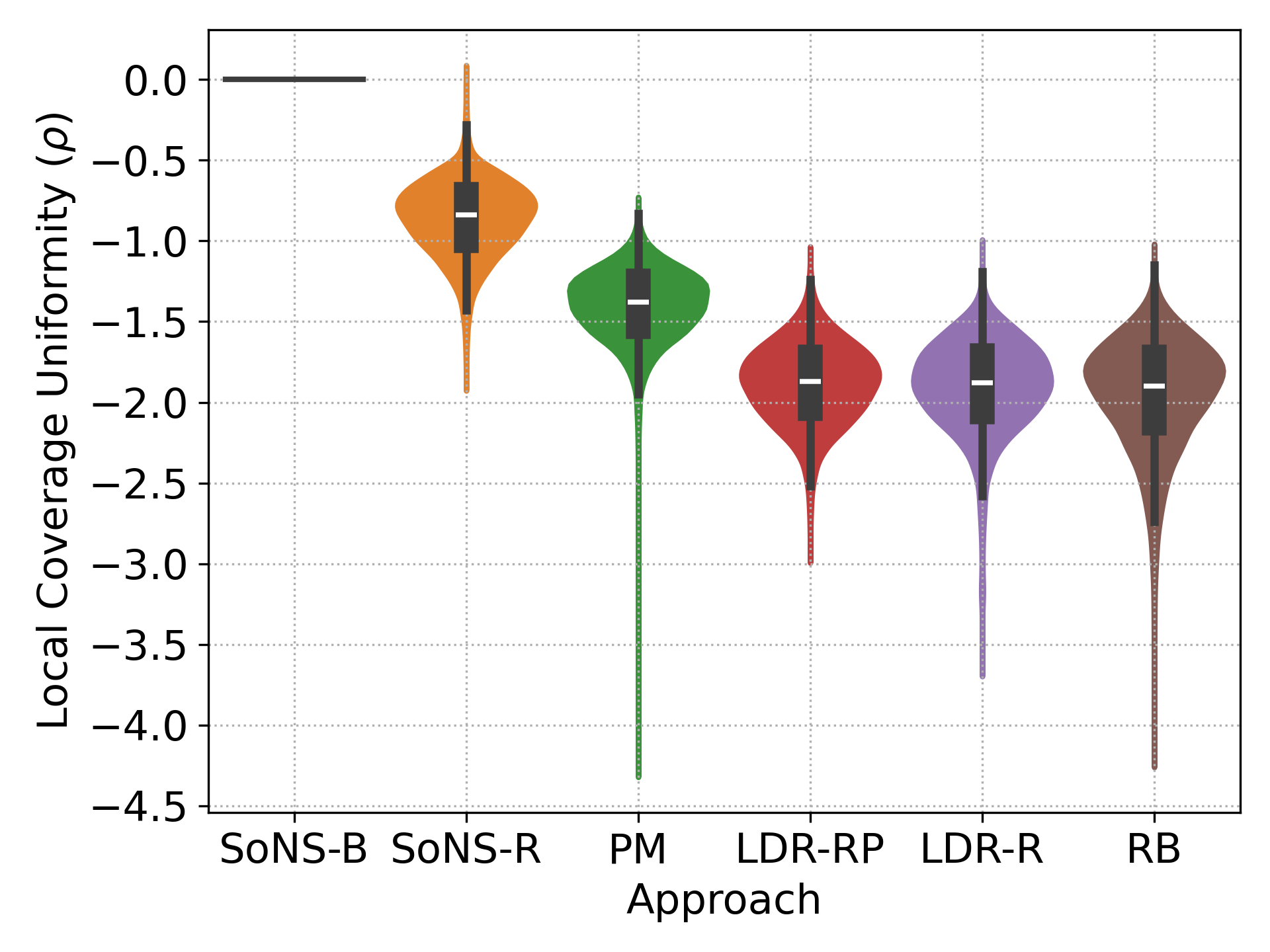}
        \label{Fig: Local Uniformity}
    }
    \caption{Coverage performance results. LDR-R and LDR-RP represent LDR-Random and LDR-Repulsive, respectively.}
    \label{fig:three_plots}
\end{figure*}

We evaluate the six approaches based on coverage completion time (CCT), the simulation step when full coverage is achieved (Figs.\ref{Fig:Time_Violin}, \ref{Fig:Time_Line}; Table\ref{tab:performance_metrics}) and the coverage progress rate (CPR) over time (Fig.~\ref{Fig:Time_Line}). \textcolor{black}{From these, SoNS-BS provides an upper-bound baseline, achieving the best \textcolor{black}{mean} CPR and CCT (2,614 steps with no variability) under its assumptions.}

The mean CPR for \textcolor{black}{SoNS}-RW is nearly the same PM but slightly worse than RB and LDR (Fig.\ref{Fig:Time_Line}). Its longer runs likely impact CPR negatively, but its mean CCT (10,194.467 steps) is significantly lower than all decentralized approaches (Fig.\ref{Fig:Time_Violin}, Table~\ref{tab:performance_metrics}). Most \textcolor{black}{SoNS}-RW runs achieve a lower CCT than decentralized approaches (Fig.~\ref{Fig:Time_Violin}).

For the benchmark decentralized approaches, UAVs struggle to locate the remaining unvisited cells as time progresses (Fig.\ref{Fig:Time_Line}). RB achieves the highest mean CPR and the lowest mean CCT (15,420.000 steps) (Figs.\ref{Fig:Time_Line}, \ref{Fig:Time_Violin}; Table~\ref{tab:performance_metrics}). The \textcolor{black}{mean} CPR of LDR-Random and LDR-Repulsive are similar and slightly better than PM. In mean CCT, PM performs the worst (22,332.800 steps), while LDR approaches are comparable, with LDR-Repulsive slightly outperforming LDR-Random (16,059.233 vs. 16,580.067 steps).

\begin{table}[t]
\vspace{2mm}
\caption{Mean and Standard Deviation of Performance Metrics.} 
\label{tab:performance_metrics}
\begin{center}
\renewcommand{\arraystretch}{1.0}  
\setlength{\tabcolsep}{3pt}  
\begin{tabular}{|c|c|c|c|c|c|c|}
\hline
\textbf{Approach} & \textbf{CCT} & \textbf{SD-CCT} & \textbf{TCU} & \textbf{SD-TCU} & \textbf{LCU} & \textbf{SD-LCU} \\ 
\hline
\textcolor{black}{SoNS-BS} & 2614.000 & 0 & 0 & 0 & 0 & 0 \\ \hline
\textcolor{black}{SoNS-RW} & 10194.467 & 2376.935 & -1.244 &  0.247 & -0.862 & 0.219 \\ \hline
PM & 22332.800 & 3961.065 &  -1.566 & 0.166 & -1.397 & 0.204 \\ \hline
LDR-RP & 16059.233 & 1555.650 & -2.0241 & 0.121 & -1.882 & 0.236 \\ \hline
LDR-R & 16580.067 & 2220.828 & -2.030 & 0.168 & -1.899 & 0.269 \\ \hline
RB & 15420.000 & 3581.138 & -2.093 & 0.284 & -1.950 & 0.327  \\ \hline
\end{tabular}
\end{center}
\end{table}

\subsection{TOTAL COVERAGE UNIFORMITY AT THE COVERAGE COMPLETION TIME (TCU)}
Coverage uniformity is defined as a measure of the variability in the number of visits across all cells in the environment. An optimal coverage strategy would have no variability, meaning all cells are visited an equal number of times. For each simulation run, let  ${v}_i \in \mathbf{v}$ represent the total number of visits to cell $i$. The coverage uniformity $\rho$ is calculated as the negative mean absolute deviation of $\mathbf{v}$ from its median, given by:

\begin{equation}
\label{eq:uniformity}
\rho = -\frac{{\sum_{i=1}^{c} {|v_i - M(\mathbf{v})|}}}{c},
\end{equation}
where $c$ is the number of cells and $M(\mathbf{v})$ is the median of $\mathbf{v}$. A higher value of $\rho$ indicates greater uniformity, with the ideal case being $\rho = 0$, where every cell is visited equally.

\textcolor{black}{As seen in Fig.\ref{Fig: Total Uniformity} and Table\ref{tab:performance_metrics}, SoNS-BS achieves the best possible TCU (mean $\rho = 0$, no variability).}

\textcolor{black}{Among random walk-based approaches, SoNS-RW achieves the highest mean TCU (-1.244), outperforming all decentralized approaches. PM follows with -1.566, making it the most uniform decentralized approach. LDR-Repulsive, LDR-Random, and RB trail behind, with mean TCU values of -2.024, -2.030, and -2.093, respectively. LDR-Repulsive shows a marginally better performance compared to the others, while RB exhibits the lowest TCU.}

\subsection{LOCAL COVERAGE UNIFORMITY AT THE COVERAGE COMPLETION TIME (LCU)}
We assess the coverage uniformity within each 10\,m\,x\,10\,m region of the environment at coverage completion using the same uniformity equation~\ref{eq:uniformity}. This metric provides insight into how evenly coverage is achieved across smaller areas within the environment.

As seen in Fig.~\ref{Fig: Local Uniformity} and Table~\ref{tab:performance_metrics}, SoNS-BS, serving as an upper-bound baseline, achieves a mean LCU of 0 with no variability, indicating perfect uniformity in each local area.

Among random walk-based approaches, SoNS-RW achieves the highest mean LCU (-0.862), significantly outperforming decentralized approaches. PM follows with -1.397, making it the best decentralized approach in this metric. LDR-Repulsive, LDR-Random, and RB yield mean LCU values of -1.882, -1.899, and -1.950, respectively, with LDR approaches slightly better than RB.

A notable observation is the higher ratio of mean TCU to mean LCU for SoNS-RW ($\approx1.44$) compared to decentralized approaches ($\approx1.12$ for PM, $\approx1.07$ for others). This higher ratio suggests a greater improvement in local uniformity relative to total uniformity for SoNS-RW, highlighting the positive impact of the SoNS framework on enhancing local uniformity.

\section{DISCUSSIONS AND CONCLUSIONS}
\textcolor{black}{In this study, we evaluated six approaches for multi-UAV sweep coverage in unknown GNSS-denied convex environments. SoNS-BS serves as an ideal baseline, providing an upper-bound performance reference under additional assumptions. Its structured line formation and back-and-forth motion result in a deterministic approach, ensuring each cell is visited exactly once and achieving perfect uniformity with no variability.}

\textcolor{black}{Among the random walk-based methods, \textcolor{black}{SoNS}-RW outperforms the decentralized approaches across all metrics.} It achieves the best TCU and LCU, with a higher ratio of mean TCU to mean LCU compared to other approaches. This indicates that \textcolor{black}{SoNS}-RW is particularly effective in distributing coverage more evenly across smaller areas, which could be advantageous in scenarios where local uniformity is crucial. Although \textcolor{black}{SoNS}-RW does not outperform in terms of variability in \textcolor{black}{all metrics}, \textcolor{black}{its mean}, maximum, and minimum values \textcolor{black}{for each metric} are significantly better than those of the decentralized approaches, indicating a much faster and more reliable approach overall.

PM is the most uniform decentralized approach, achieving good TCU and LCU. However, this comes at the cost of time; PM has the slowest CCT among all approaches. While its effectiveness in achieving uniform coverage is well-established, its slower progress highlights the uniformity-speed trade-off in decentralized coverage strategies.

Among the decentralized approaches, RB demonstrates the lowest mean CCT and the highest mean CPR. However, RB has the lowest consistency (highest variability) across all three metrics, which can lead to less reliable performance in varied environments. LDR-Repulsive and LDR-Random show comparable results in terms of mean TCU and LCU, but LDR-Repulsive has the lowest variability, making it the most consistent decentralized approach. However, this consistency requires the assumption that UAVs 
can accurately estimate the relative position of their neighbors upon receiving a message, which may not be realistic.

While \textcolor{black}{SoNS}-RW has a slightly lower mean CPR compared to the fastest decentralized approach, RB, several factors contribute to this outcome. In our \textcolor{black}{SoNS} setup, only 20 out of 25 UAVs are active as sampling UAVs, and \textcolor{black}{SoNS}-RW allows UAVs to cross the environment's borders to perform turns and select new directions. Consequently, for a significant amount of time, a large portion of the \textcolor{black}{SoNS} may fly outside the arena, resulting in fewer active sampling UAVs within the environment and occasionally leading to a lower CPR compared to the benchmark approaches. Future work could explore adaptive \textcolor{black}{SoNS}-based random walk strategies that dynamically adjust the formation upon detecting a boundary, combining advanced shifts and curved back-and-forth sweeps while ensuring all UAVs remain within the environment, thereby improving overall performance.

Reducing density using LDR approaches results in higher consistency in performance across all metrics without necessarily improving the mean values for CCT, TCU, and LCU, and often at the cost of time compared to RB. This finding is influenced by the resolution of the decomposition (cell size). If larger cells were considered or if cells were marked as visited when within the UAV's field of view rather than being physically entered, density reduction would likely improve CCT, TCU, and LCU. However, in our setup, UAVs might miss visiting an unvisited cell due to their reactions, which can lead to increased coverage time. An adaptive density reduction strategy could be another avenue for future exploration to further improve performance.


\textcolor{black}{In summary, SoNS-BS serves as an upper-bound baseline, demonstrating the best performance under its assumptions. SoNS-RW} shows better performance than decentralized approaches in all three metrics and is particularly advantageous where local uniformity is a priority. PM offers the best uniformity among the decentralized approaches, but at a significant time cost, and \textcolor{black}{RB is} fast but inconsistent \textcolor{black}{and has} low uniformity. LDR approaches improve consistency.













\begin{thebibliography}{10}

\bibitem{senanayake2016search}
M.~Senanayake, I.~Senthooran, and et~al., ``Search and tracking algorithms for swarms of robots: A survey,'' {\em Robotics and Autonomous Systems}, vol.~75, no.~Part B, pp.~422--434, 2016.

\bibitem{pham2017distributed}
H.~X. Pham, H.~M. La, D.~Feil-Seifer, and M.~Deans, ``A distributed control framework for a team of unmanned aerial vehicles for dynamic wildfire tracking,'' in {\em 2017 IEEE/RSJ international conference on intelligent robots and systems (IROS)}, pp.~6648--6653, IEEE, 2017.

\bibitem{casbeer2006cooperative}
D.~W. Casbeer, D.~B. Kingston, R.~W. Beard, and T.~W. McLain, ``Cooperative forest fire surveillance using a team of small unmanned air vehicles,'' {\em International journal of systems science}, vol.~37, no.~6, pp.~351--360, 2006.

\bibitem{lin2018topology}
Z.~Lin and H.~H. Liu, ``Topology-based distributed optimization for multi-uav cooperative wildfire monitoring,'' {\em Optimal Control Applications and Methods}, vol.~39, no.~4, pp.~1530--1548, 2018.

\bibitem{basilico2015deploying}
N.~Basilico and S.~Carpin, ``Deploying teams of heterogeneous uavs in cooperative two-level surveillance missions,'' in {\em 2015 IEEE/RSJ International Conference on Intelligent Robots and Systems (IROS)}, pp.~610--615, IEEE, 2015.

\bibitem{kegeleirs2019random}
M.~Kegeleirs, D.~Garz{\'o}n~Ramos, and M.~Birattari, ``Random walk exploration for swarm mapping,'' in {\em Annual conference towards autonomous robotic systems}, pp.~211--222, Springer, 2019.

\bibitem{dimidov2016random}
C.~Dimidov, G.~Oriolo, and V.~Trianni, ``Random walks in swarm robotics: an experiment with kilobots,'' in {\em International conference on swarm intelligence}, pp.~185--196, Springer, 2016.

\bibitem{tan2021comprehensive}
C.~S. Tan, R.~Mohd-Mokhtar, and M.~R. Arshad, ``A comprehensive review of coverage path planning in robotics using classical and heuristic algorithms,'' {\em IEEE Access}, vol.~9, pp.~119310--119342, 2021.

\bibitem{jamshidpey2020multi}
A.~Jamshidpey, W.~Zhu, M.~Wahby, M.~Allwright, M.~K. Heinrich, and M.~Dorigo, ``Multi-robot coverage using self-organized networks for central coordination,'' in {\em International Conference on Swarm Intelligence}, pp.~216--228, Springer, 2020.

\bibitem{zhu2020formation}
W.~Zhu, M.~Allwright, M.~K. Heinrich, S.~O{\u{g}}uz, A.~L. Christensen, and M.~Dorigo, ``Formation control of uavs and mobile robots using self-organized communication topologies,'' in {\em International conference on swarm intelligence}, pp.~306--314, Springer, 2020.

\bibitem{zhu2024self}
W.~Zhu, S.~Oguz, M.~K. Heinrich, M.~Allwright, M.~Wahby, A.~L. Christensen, E.~Garone, and M.~Dorigo, ``Self-organizing nervous systems for robot swarms,'' {\em arXiv preprint arXiv:2401.13103}, 2024.

\bibitem{liu2018survey}
Y.~Liu and R.~Bucknall, ``A survey of formation control and motion planning of multiple unmanned vehicles,'' {\em Robotica}, vol.~36, no.~7, pp.~1019--1047, 2018.

\bibitem{jamshidpey2023reducing}
A.~Jamshidpey, M.~Dorigo, and M.~K. Heinrich, ``Reducing uncertainty in collective perception using self-organizing hierarchy,'' {\em Intelligent Computing}, vol.~2, p.~0044, 2023.

\bibitem{choset1998coverage}
H.~Choset and P.~Pignon, ``Coverage path planning: The boustrophedon cellular decomposition,'' in {\em Field and service robotics}, pp.~203--209, Springer, 1998.

\bibitem{elhabyan2019coverage}
R.~Elhabyan, W.~Shi, and M.~St-Hilaire, ``Coverage protocols for wireless sensor networks: Review and future directions,'' {\em Journal of Communications and Networks}, vol.~21, no.~1, pp.~45--60, 2019.

\bibitem{gorain2014approximation}
B.~Gorain and P.~S. Mandal, ``Approximation algorithms for sweep coverage in wireless sensor networks,'' {\em Journal of parallel and Distributed Computing}, vol.~74, no.~8, pp.~2699--2707, 2014.

\bibitem{GALCERAN20131258}
E.~Galceran and M.~Carreras, ``A survey on coverage path planning for robotics,'' {\em Robotics and Autonomous Systems}, vol.~61, no.~12, pp.~1258--1276, 2013.

\bibitem{cabreira2019survey}
T.~M. Cabreira, L.~B. Brisolara, and F.~J. Paulo~R, ``Survey on coverage path planning with unmanned aerial vehicles,'' {\em Drones}, vol.~3, no.~1, p.~4, 2019.

\bibitem{10421780}
B.~Jia, Z.~Gao, J.~Jing, B.~Huang, S.~Liu, K.~Muhammad, and J.~J. P.~C. Rodrigues, ``Coverage path planning for iouavs with tiny machine learning in complex areas based on convex decomposition,'' {\em IEEE Internet of Things Journal}, vol.~11, no.~12, pp.~21103--21111, 2024.

\bibitem{8894171}
M.~Hassan and D.~Liu, ``Ppcpp: A predator–prey-based approach to adaptive coverage path planning,'' {\em IEEE Transactions on Robotics}, vol.~36, no.~1, pp.~284--301, 2020.

\bibitem{10102336}
W.~Hu, Y.~Yu, S.~Liu, C.~She, L.~Guo, B.~Vucetic, and Y.~Li, ``Multi-uav coverage path planning: A distributed online cooperation method,'' {\em IEEE Transactions on Vehicular Technology}, vol.~72, no.~9, pp.~11727--11740, 2023.

\bibitem{8286947}
J.~Song and S.~Gupta, ``$\varepsilon ^{\star }$: An online coverage path planning algorithm,'' {\em IEEE Transactions on Robotics}, vol.~34, no.~2, pp.~526--533, 2018.

\bibitem{feynman1963mainly}
R.~P. Feynman, R.~B. Leighton, and M.~Sands, ``Mainly mechanics, radiation, and heat,'' {\em (No Title)}, 1963.

\bibitem{renshaw1981correlated}
E.~Renshaw and R.~Henderson, ``The correlated random walk,'' {\em Journal of Applied Probability}, vol.~18, no.~2, pp.~403--414, 1981.

\bibitem{zaburdaev2015levy}
V.~Zaburdaev, S.~Denisov, and J.~Klafter, ``L{\'e}vy walks,'' {\em Reviews of Modern Physics}, vol.~87, no.~2, pp.~483--530, 2015.

\bibitem{pasternak2009levy}
Z.~Pasternak, F.~Bartumeus, and F.~W. Grasso, ``L{\'e}vy-taxis: a novel search strategy for finding odor plumes in turbulent flow-dominated environments,'' {\em Journal of Physics A: Mathematical and Theoretical}, vol.~42, no.~43, p.~434010, 2009.

\bibitem{comets2009billiards}
F.~Comets, S.~Popov, G.~M. Sch{\"u}tz, and M.~Vachkovskaia, ``Billiards in a general domain with random reflections,'' {\em Archive for rational mechanics and analysis}, vol.~191, no.~3, pp.~497--537, 2009.

\bibitem{dorigo1992optimization}
M.~Dorigo, ``Optimization, learning and natural algorithms,'' {\em Ph. D. Thesis, Politecnico di Milano}, 1992.

\bibitem{rosalie2018chaos}
M.~Rosalie, G.~Danoy, S.~Chaumette, and P.~Bouvry, ``Chaos-enhanced mobility models for multilevel swarms of uavs,'' {\em Swarm and evolutionary computation}, vol.~41, pp.~36--48, 2018.

\bibitem{ge2005complete}
S.~S. Ge and C.-H. Fua, ``Complete multi-robot coverage of unknown environments with minimum repeated coverage,'' in {\em Proceedings of the 2005 IEEE International Conference on Robotics and Automation}, pp.~715--720, IEEE, 2005.

\bibitem{kuiper2006mobility}
E.~Kuiper and S.~Nadjm-Tehrani, ``Mobility models for uav group reconnaissance applications,'' in {\em 2006 International Conference on Wireless and Mobile Communications (ICWMC'06)}, pp.~33--33, IEEE, 2006.

\bibitem{albani2017field}
D.~Albani, D.~Nardi, and V.~Trianni, ``Field coverage and weed mapping by uav swarms,'' in {\em 2017 IEEE/RSJ International Conference on Intelligent Robots and Systems (IROS)}, pp.~4319--4325, Ieee, 2017.

\bibitem{stevens1997aggregation}
A.~Stevens and H.~G. Othmer, ``Aggregation, blowup, and collapse: the abc's of taxis in reinforced random walks,'' {\em SIAM Journal on Applied Mathematics}, vol.~57, no.~4, pp.~1044--1081, 1997.

\bibitem{devaraju2023connectivity}
S.~Devaraju, A.~Ihler, and S.~Kumar, ``A connectivity-aware pheromone mobility model for autonomous uav networks,'' in {\em 2023 IEEE 20th Consumer Communications \& Networking Conference (CCNC)}, pp.~1--6, IEEE, 2023.

\bibitem{jamshidpey2024centralization}
A.~Jamshidpey, M.~Wahby, M.~K. Heinrich, M.~Allwright, W.~Zhu, and M.~Dorigo, ``Centralization vs. decentralization in multi-robot coverage: Ground robots under uav supervision,'' {\em arXiv preprint arXiv:2408.06553}, 2024.

\bibitem{bayert2019robotic}
J.~Bayert and S.~Khorbotly, ``Robotic swarm dispersion using gradient descent algorithm,'' in {\em 2019 IEEE International Symposium on Robotic and Sensors Environments (ROSE)}, pp.~1--6, IEEE, 2019.

\bibitem{vijay2017received}
M.~Vijay, M.~Kuber, and K.~Sivayazi, ``Received signal strength based dispersion of swarm of autonomous ground vehicles,'' in {\em 2017 2nd IEEE International Conference on Recent Trends in Electronics, Information \& Communication Technology (RTEICT)}, pp.~52--57, IEEE, 2017.

\bibitem{beal2013superdiffusive}
J.~Beal, ``Superdiffusive dispersion and mixing of swarms with reactive levy walks,'' in {\em 2013 IEEE 7th International Conference on Self-Adaptive and Self-Organizing Systems}, pp.~141--148, IEEE, 2013.

\bibitem{khaluf2018collective}
Y.~Khaluf, S.~Van~Havermaet, and P.~Simoens, ``Collective l{\'e}vy walk for efficient exploration in unknown environments,'' in {\em Artificial Intelligence: Methodology, Systems, and Applications: 18th International Conference, AIMSA 2018, Varna, Bulgaria, September 12--14, 2018, Proceedings 18}, pp.~260--264, Springer, 2018.

\bibitem{pinciroli2012argos}
C.~Pinciroli, V.~Trianni, R.~O’Grady, G.~Pini, A.~Brutschy, M.~Brambilla, N.~Mathews, E.~Ferrante, G.~Di~Caro, F.~Ducatelle, {\em et~al.}, ``Argos: a modular, parallel, multi-engine simulator for multi-robot systems,'' {\em Swarm intelligence}, vol.~6, pp.~271--295, 2012.

\bibitem{allwright2018argos}
M.~Allwright, N.~Bhalla, C.~Pinciroli, and M.~Dorigo, ``Argos plug-ins for experiments in autonomous construction,'' tech. rep., Tech. Rep. TR/IRIDIA/2018-007, IRIDIA, Universit{\'e} Libre de Bruxelles~…, 2018.

\bibitem{allwright2018simulating}
M.~Allwright, N.~Bhalla, C.~Pinciroli, and M.~Dorigo, ``Simulating multi-robot construction in argos,'' in {\em International Conference on Swarm Intelligence}, pp.~188--200, Springer, 2018.

\end{thebibliography}
\end{document}